\documentclass[letterpaper,twocolumn,10pt]{article}
\usepackage{usenix-2020-09}

\usepackage{tikz}
\usepackage{amsmath}
\usepackage{enumitem}
\usepackage{float}
\usepackage{parskip}
\usepackage{tabularx}
\usepackage{xcolor}
\usepackage{multirow} 
\usepackage{graphicx}
\usepackage{subcaption}
\usepackage[linesnumbered]{algorithm2e}
\SetKwComment{Comment}{/* }{ */}
\RestyleAlgo{ruled}

\usepackage{filecontents}
\usepackage{amsmath}

\newcommand{\eg}{{e.g.}}
\newcommand{\ie}{{i.e.}}
\newcommand{\projectname}{Optimus}
\definecolor{forwardcolor}{RGB}{0,0,255} 
\definecolor{backwardcolor}{RGB}{128,0,128} 

\newcommand{\revise}[1]{\textcolor{black}{#1}}

\usepackage[scaled]{inconsolata}   

\begin{document}
\pagestyle{empty}

\date{}

\title{\Large \bf \projectname: Accelerating Large-Scale Multi-Modal LLM Training by Bubble Exploitation}

\newif\ifshowauthors
\showauthorstrue 

\author{
\ifshowauthors
    {\rm Weiqi Feng$^1$\footnotemark[1], Yangrui Chen$^2$, Shaoyu Wang$^3$\footnotemark[1], Yanghua Peng$^2$, Haibin Lin$^2$ and Minlan Yu$^1$} \\
    $^1$Harvard University, $^2$Bytedance, 
    $^3$University of Southern California
\else
    Paper \#174, 12 pages 
\fi
} 

\maketitle

\begingroup
\renewcommand\thefootnote{*}
\footnotetext{Work done during internship at Bytedance.}
\endgroup

\begin{abstract}

Multimodal large language models (MLLMs) have extended the success of large language models (LLMs) to multiple data types, such as image, text and audio, achieving significant performance in various domains, including multimodal translation, visual question answering and content generation. Nonetheless, existing systems are inefficient to train MLLMs due to substantial GPU bubbles caused by the heterogeneous modality models and complex data dependencies in 3D parallelism. 
This paper proposes {\projectname}, a distributed MLLM training system that reduces end-to-end MLLM training time. 
{\projectname} is based on our principled analysis that scheduling the encoder computation within the LLM bubbles can reduce bubbles in MLLM training. 
To enable scheduling encoder computation for all GPUs, {\projectname} searches for separate parallel plans for the encoder and LLM, and adopts a bubble scheduling algorithm to exploit LLM bubbles without breaking the original data dependencies in the MLLM model architecture. We further decompose the encoder layer computation into a series of kernels and analyze the common bubble pattern of 3D parallelism to carefully optimize the sub-millisecond bubble scheduling, minimizing the overall training time. 
Our experiments in a production cluster show that {\projectname} accelerates MLLM training by 20.5\%-21.3\% with ViT-22B and GPT-175B model over 3072 GPUs compared to baselines.
    
\end{abstract}

\vspace{-10pt}
\section{Introduction}
\label{sec:intro}

Multimodal Large Language Models (MLLMs) build upon the advancements of Large Language Models (LLMs) by extending their capabilities to comprehend and generate content across multiple modalities, such as text, images, and audio.
Notable MLLMs, including GPT-4V~\cite{gpt4v}, Google Gemini~\cite{team2023gemini}, Grok-1.5 Vision~\cite{grok1.5v} and LLava~\cite{liu2023llava}, have achieved remarkable progress in domains like visual question answering~\cite{agrawal2016vqa,marino2019ok}, multimodal translation~\cite{sulubacak2020multimodal,yao2020multimodal}, and content generation and understanding~\cite{team2023gemini,gpt4v,zhu2023minigpt}. 
The substantial computational demands of MLLMs underscore the urgent need to enhance training performance to fully harness their capabilities.

Multimodal large language models (MLLMs) typically integrate multiple encoders, each specialized for processing a specific modality, alongside a substantial language model component. Multimodal data are input into their respective encoders, and the resulting outputs are combined to form the input for the language model.

The multimodal encoders and the language model exhibit significant differences in functionality, architecture, and input data sizes, leading to varied resource demands. However, existing distributed training systems are mainly designed for sequential unimodal (\eg, MegaScale~\cite{jiang2024megascale}, Megatron-LM~\cite{narayanan2021efficient}, Chimera~\cite{li2021chimera}), and fall short in training MLLMs. For example, when training a large MLLM containing several hundred billion parameters using Megatron-LM on over 3,000 GPUs, more than 40\% of GPU cycles remain idle. Upon analyzing typical MLLM training tasks, we identified two critical issues: (1) Communication overhead in 3D parallelism is extensive and frequent, resulting in significant GPU idle times; (2) The pipeline stages of MLLM are imbalanced, and the data dependency between adjacent pipeline stages results in considerable data waiting time. Existing solutions can be categorized into two groups: (1) optimizing LLM, \eg Megatron-LM and Zero-bubble pipeline\cite{qi2023zero}; (2) optimizing multimodal encoders, \eg DistMM\cite{Huang2024}. Nonetheless, none of the existing works consider LLM and encoders simultaneously. As demonstrated in Section~\ref{subsec:bubble_in_mllm}, around 48\% of GPU cycles are wasted in our internal large-scale MLLM training task. 

In this paper, we introduce {\projectname}, a distributed MLLM training system that enables scheduling encoder computations within idle periods — referred to as "LLM bubbles" — to achieve efficient 3D parallelism. Scheduling encoder computations within these LLM bubbles using existing training frameworks is challenging for three reasons.

\textit{First}, existing training frameworks, \eg, Megatron-LM~\cite{narayanan2021efficient}, MegaScale~\cite{jiang2024megascale}, and zero-bubble pipeline~\cite{qi2023zero}, employ unified parallel strategies to MLLM models, distributing encoder and LLM layers across different GPUs. As a result, most GPUs contain only LLM model states and are incapable of performing encoder computations during LLM bubbles.
In contrast, our approach utilizes separate parallelization plans for encoders and LLMs to colocate both encoder and LLM model states on each GPU. We systematically enumerate potential 3D parallelism plans for the encoder and eliminate those that violate GPU memory constraints.

\textit{Second}, complex data dependencies inherent in MLLM constrain the scheduling of encoder computations within LLM bubbles. These dependencies include those related to synchronous training iterations and internal dependencies within the encoder itself, as discussed in Section \ref{subsec:challenge}. The most intricate dependency is the microbatch-level data dependency between the encoder and LLM, which requires that the encoder completes its forward pass before the LLM begins its forward pass for each microbatch, and that the encoder's backward pass commences only after the LLM has completed its backward pass for each microbatch. To manage these dependencies, we employ a two-stage dependency management strategy: local scheduling to address the first two types of dependencies and global ordering to handle the encoder-LLM microbatch-level dependencies.

\textit{Third}, the LLM bubble duration varies significantly, ranging from sub-millisecond intervals to several hundred milliseconds, posing a considerable challenge for bubble reduction. Existing frameworks~\cite{narayanan2021efficient, qi2023zero, li2021chimera} schedule computations at the layer level, and sub-millisecond bubbles are too brief to complete even a single encoder layer forward or backward. To address this, we decompose encoder layer computations into sequences of kernels, enabling the effective utilization of these brief bubbles. Furthermore, we analyze common patterns in LLM bubbles and optimize the scheduling by interleaving encoder kernel computations with LLM computations, thereby minimizing the overall training time.

{\projectname} was implemented based on Megatron-LM, incorporating the aforementioned design principles. Comprehensive evaluations were conducted using several representative MLLM models. The results are promising - {\projectname} outperforms state-of-the-art baselines by 20.3\% on average and {\projectname} also scales well with the size of models and GPUs. 
Our experiments in a production cluster show that {\projectname} accelerates MLLM training by 20.5\%-21.3\% with ViT-22B and GPT-175B models over 3072 GPUs compared to baselines.

\vspace{-10pt}

\section{Background}

\subsection{Multimodal LLM Characteristics}
\label{subsec:mllm_background}

Multimodal LLMs are increasingly important. These models build upon the foundational principles of LLMs by integrating advanced natural language processing methodologies while extending their scope to encompass diverse data modalities. GPT-4\cite{gpt4v} exemplifies such an advancement, enhancing the capabilities of its predecessors to include multimodal understanding and generation. It demonstrates human-level performance across various benchmark tests involving both image and text inputs.

\begin{figure}[htp]
\vspace{-10pt}
    \centering
    \includegraphics[scale=0.8]{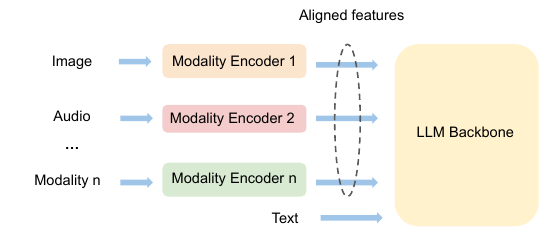}
    \caption{Multimodal model architecture.}
    \label{fig:mllm_arch}
    \vspace{-10pt}
\end{figure}

A multimodal large language model (MLLM) typically comprises three primary components: one or more modality encoders, input projectors, and a large language model backbone~\cite{zhang2024mm}. The Modality Encoders process inputs from non-textual modalities into their respective feature representations, while the input projector aligns these features with the text feature space. Subsequently, the LLM backbone utilizes the aligned multimodal and textual features as input. Figure \ref{fig:mllm_arch} illustrates the architecture of the MLLM. The input projector is excluded from further discussion due to its relatively minimal computational demands compared to the encoder and the LLM backbone, as detailed in Llava~\cite{liu2023llava}. For this analysis, the input projector is considered the final layer of the modality encoder.

Different from homogeneous LLM architecture, multimodal LLM has the following unique characteristics.

{\bf Dominant Model Size of LLM Backbone:} In multimodal LLMs, the LLM backbone has a significantly larger number of parameters compared to other components such as encoders and projectors. For instance, Flamingo \cite{alayrac2022flamingo} boasts a total of 80 billion parameters, with its LLM backbone alone comprising 70 billion parameters.

{\bf Dependency between Encoders and LLM Backbone:} In MLLM training, there are two types of data dependencies between encoders and LLM. During the forward pass, encoders must complete the generation of encoded features before the LLM backbone can proceed with forwarding. Conversely, in the backward pass, the LLM backbone calculates gradients before the encoders initiate the backward pass.

\subsection{Bubbles in MLLM Training}
\label{subsec:bubble_in_mllm}
Existing LLM pipeline optimizations lack model-agnostic properties and are inadequate for MLLM training tasks. 
In our internal large-scale MLLM training tasks with ViT encoder and GPT backbone (over 100B parameters), we employed Megatron-LM across more than 3,000 NVIDIA GPUs and observed over 48\% GPU cycle idleness despite utilizing multiple SOTA techniques, including MegaScale~\cite{jiang2024megascale}, Zero Bubble Pipeline~\cite{qi2023zero}, fine-grained communication-computation overlapping~\cite{wang2022overlap}. We analyzed the profiled timeline to identify and investigate the occurrences of GPU idleness (\ie, bubbles).
Table \ref{tab:production_data} shows the total duration and percentage of average training step time (5.12s) occupied by different types of bubbles.

\begin{table}[htp]
    \centering
\begin{tabular}{l|c|c}
    \hline
     Bubble types & Percentage & Total time (s) \\
    \hline
    DP bubble (all-gather)  & 3.3\% & 0.167\\
    DP bubble (reduce-scatter)  & 8.9\%  & 0.458\\
    \hline
    PP bubbles (warmup)  & 5.0\% & 0.291 \\
    PP bubbles (cooldown) & 9.2\% & 0.471 \\
    PP bubbles (other) & 8.7\%  & 0.445 \\
    \hline
    TP bubble & 11.2\% & 0.585\\
    \hline
\end{tabular}
    \caption{Total time and percentage of average training step time (5.12s) occupied by different types of bubbles.}
    \label{tab:production_data}
\end{table}

These bubbles can be classified into three categories based on their underlying causes.

\begin{figure}[htp]
    \centering
    \includegraphics[width=0.48\textwidth]{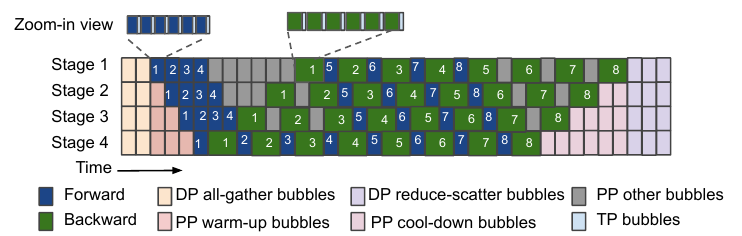}
    \caption{Timeline illustration of MLLM training showing different categories of bubbles (simplified based on the Megatron-LM 1F1B schedule \cite{narayanan2021efficient}).} 
    \label{fig:bubble_existing}
\end{figure}

\textbf{(1) Communication in Data Parallelism (DP).} Data parallelism requires communication to aggregate gradients, leading to GPU idle time during these communications.
Specifically, MegaScale~\cite{jiang2024megascale} and Megatron-LM~\cite{DBLP:journals/corr/abs-1909-08053} employ a distributed optimizer, similar to $P_{os+g}$ in ZeRO \cite{rajbhandari2020zero}, to save memory for large model training, which performs two collective communications: \textit{all-gather} and \textit{reduce-scatter}.  At the start of each training step, an all-gather operation collects updated parameters from all DP ranks, resulting in a DP all-gather bubble that occupies 3.3\% of the training time. At the end of the training step, reduce-scatter is performed to aggregate gradients, leading to a DP reduce-scatter bubble that consumes 8.9\% of the training time\footnote{\revise{The reduce-scatter bubble is larger than the all-gather bubble because it involves higher communication volume (FP32 gradients vs. BF16 parameters) and suffers from synchronization delays due to straggling GPUs at the end of the training step.}}.
It should be noted that overlapping optimization in data parallelism proposed in Megascale~\cite{jiang2024megascale} has already been applied; however, the aforementioned DP communications for the first model chunk can not be hidden due to the nature of synchronous training~\cite{jiang2024megascale}. Figure \ref{fig:bubble_existing} illustrates DP bubbles arising from all-gather and reduce-scatter operations at the start and conclusion of each training step.

\textbf{(2) Dependency in Pipeline Parallelism (PP)}. 
Despite applying pipeline send-receive overlap optimization from Megascale \cite{jiang2024megascale}, pipeline bubbles persist due to the inherent data dependencies between stages during the forward and backward passes. Importantly, the Zero Bubble Pipeline method cannot eliminate pipeline bubbles in MLLM training, owing to the required changes in the optimizer \cite{qi2023zero} (see discussions in \S\ref{sec:related_works}).
Figure \ref{fig:bubble_existing} illustrates the MLLM training pipeline schedule, comprising three phases: warm-up (forward only), steady (one forward and one backward), and cool-down (backward only).  
Throughout pipeline training, three types of bubbles arise:
\vspace{-5pt}
\begin{enumerate}[label=\textbullet,itemsep=-5pt]
\item PP warm-up bubbles occur at all stages except the first one due to the forward dependency of the first forward pass, averaging 5.0\% of the training time.

\item PP cool-down bubbles occur at all stages except the first one due to the backward dependency of the final backward pass, averaging 9.2\% of the training time.

\item Other PP bubbles manifest in all stages except the last one due to dependencies of other forward and backward passes, occupying 8.7\% of training time. For instance, PP bubbles emerge immediately after the PP warm-up phase due to the backward dependency of the initial backward pass. Additionally, in cases where pipeline stages are imbalanced due to the heterogeneity of MLLM models, extra pipeline bubbles not depicted in Figure \ref{fig:bubble_existing} occur.
\end{enumerate}

\begin{figure}[htp]
    \vspace{-10pt}
     \centering
     \includegraphics[width=0.45\textwidth]{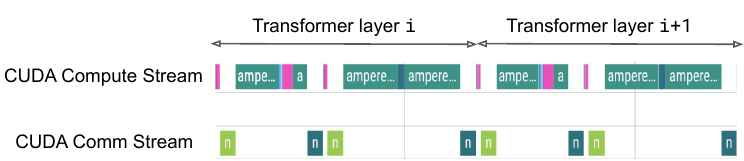}
     \caption{Zoom-in view of TP bubbles during two LLM layers forward.}
     \label{fig:profiled_tp_bubble}
    \vspace{-10pt}
 \end{figure}

\textbf{(3) Communications in Tensor Parallelism (TP)}. Tensor parallelism involves partitioning individual layers across multiple GPUs, necessitating communication during forward and backward passes to synchronize between GPUs. In Megatron-LM, each forward or backward pass of a transformer layer requires two all-gather and two reduce-scatter kernels \cite{korthikanti2023reducing}. Figure \ref{fig:profiled_tp_bubble} provides a detailed view of CUDA computation and communication kernels during two GPT-175B~\cite{brown2020language} layer forward passes. Green kernels represent all-gather communications in the CUDA communication stream, while blue kernels denote reduce-scatter communications. The compute stream idles during these communications. Typically, these TP bubbles last for sub-millisecond durations, averaging around 300 $\mu$s. However, during MLLM training, there are thousands of TP bubbles, collectively accounting for 11.2\% of the training time.

\subsection{Challenges}
\label{subsec:challenge}
To minimize bubbles in MLLM training, we aim to exploit the distinct dual-component structure of MLLM, which consists of encoders and the LLM backbone. We make two key observations.
First, the majority of bubbles during MLLM training occur during the forward and backward passes of the LLM backbone, with around 90\% arising from LLM communication based on production data. Second, the encoders require fewer computational operations (FLOPs) than the LLM backbone due to their smaller parameter size\cite{bai2023qwenvl, chen2023minigptv2, liu2023llava, liu2024llavanext, driess2023palme}.

Based on these insights, we propose \textbf{scheduling encoder computations in LLM bubbles} which occur during LLM communication to minimize bubbles throughout the MLLM training process. We identify three primary challenges of scheduling encoder computation to LLM bubbles. 

\begin{figure}[htp]
    \centering
 \includegraphics[scale=0.8]{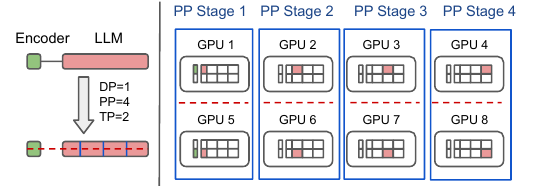}
    \caption{Only GPUs in pipeline stage 1 have both encoder and LLM model states.}
    \label{fig:model_shards_a}
\vspace{-10pt}
\end{figure}

\textbf{Challenge 1: Only a limited number of GPUs have both encoder and LLM model states.}
Current training systems \cite{zheng2022alpa,narayanan2021efficient} employ pipeline parallelism to distribute the MLLM as a single pipeline across multiple GPUs. Due to the dependency between the encoder and LLM, encoder layers are assigned to initial pipeline stages, while LLM layers are assigned to later pipeline stages. Consequently, only one pipeline stage typically contains both the encoder and LLM layers. To illustrate, Figure \ref{fig:model_shards_a} demonstrates the use of 3D parallelism (DP=1, PP=4, TP=2) to parallelize MLLM across 8 GPUs, where only 2 GPUs in pipeline stage 1 hold both the encoder and LLM model states. The remaining 6 GPUs cannot execute encoder computations during LLM bubbles because they lack encoder model states.

\textbf{Challenge 2: Complex Dependencies in MLLM Training}. The complex dependencies inherent in MLLM training pose significant challenges when scheduling encoder computation within LLM bubbles. First, in synchronous training, the utilization of LLM bubbles is restricted to performing encoder computations solely within the current training iteration (\textit{iteration dependency}). Second, dependencies within the encoder pipeline require that the forward computation of the current encoder pipeline stage $i$ is scheduled only after the completion of the previous encoder stages, and the backward computation can be scheduled only after the subsequent encoder stage concludes. Lastly, the encoder-LLM dependency imposes microbatch-level constraints: the encoder must complete the forward pass of microbatch $i$ before the LLM pipeline can initiate the forward pass of the same microbatch. Similarly, the encoder can begin the backward pass of microbatch $i$ only after the LLM pipeline has completed its backward pass of microbatch $i$.

\textbf{Challenge 3: Sub-millisecond LLM bubbles.} Existing frameworks like MegaScale\cite{jiang2024megascale} and Megatron-LM\cite{narayanan2021efficient} typically schedule at layer level. However, bubbles in the LLM exhibit a wide range of durations, spanning from sub-milliseconds~(TP bubbles) to hundreds of milliseconds~(DP bubbles). For instance, TP bubbles shown in Figure \ref{fig:profiled_tp_bubble} average around 300$\mu$s across different LLM layers during forward and backward passes. This duration is insufficient to complete even a single encoder layer forward or backward. To illustrate, a single ViT-22B layer typically requires around 1.4 milliseconds to complete forward propagation and 2.0 milliseconds to complete backward propagation. 

\vspace{-10pt}
\section{Design Decisions and System Overview}

We discuss the core design decisions that drive \projectname{} design and provide an overview of \projectname{}. The next section discusses the detailed design.

\vspace{-10pt}
\subsection{Design Decisions}
\label{subsec:insights_challenges}

\begin{figure}[!t]
    \centering
 \includegraphics[scale=0.8]{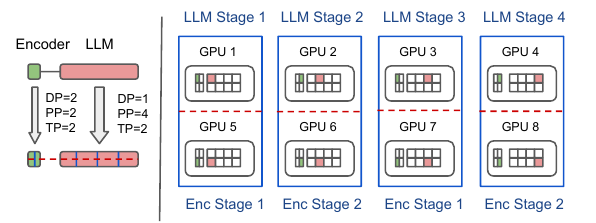}
    \caption{All GPUs both hold encoder and LLM model states when giving encoder and LLM separate parallel plans.}
    \label{fig:model_shards_b}
    \vspace{-10pt}
\end{figure}

\textbf{Design decision 1: Colocate encoders and LLM with separate parallelism.} To ensure that each GPU possesses both encoder and LLM model states, we propose assigning separate parallel plans to encoders and LLMs across all GPUs. This strategy is illustrated in Figure \ref{fig:model_shards_b}, where using parallel plan (DP=2, PP=2, TP=2) for encoders and (DP=1, PP=4, TP=2) for LLM. Each GPU retains both encoder and LLM model states, and then it becomes feasible for all GPUs to execute encoder computations during LLM bubbles. 
Note that colocating both the encoder and LLM states may require more GPU memory, and we analyze the memory overhead in Section \ref{subsec:algo_memory}.

\begin{figure}[htp]
\vspace{-10pt}
    \centering
    \includegraphics[width=0.48\textwidth]{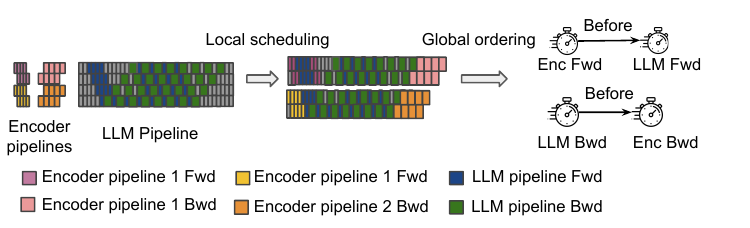}
    \caption{Solve complex dependencies in MLLM training through local scheduling and global ordering.}
    \label{fig:design_2}
\end{figure}

\textbf{Design decision 2: Dual-Stage Dependency Management.} 
 We use two stages to handle complex dependencies in MLLM training: local scheduling and global ordering.
Each encoder pipeline undergoes local scheduling, which schedules encoder computations with available LLM bubbles, adhering to the iteration-dependency and encoder-internal dependencies. Global ordering ensures microbatch-level dependency between encoders and LLM by sequencing the encoder's ending times forward and the encoder's starting times backward across microbatches. 
This involves comparing timestamps to verify encoder-LLM dependency compliance. As shown in Figure \ref{fig:design_2}, local scheduling is applied independently to two encoder pipelines, maintaining iteration dependency and encoder-internal dependency. In global ordering, timestamps across all microbatches (totaling 8) are checked to confirm that encoder-LLM dependencies are met.

\begin{figure}[h]
\vspace{-10pt}
    \centering
    \includegraphics[scale=0.7]{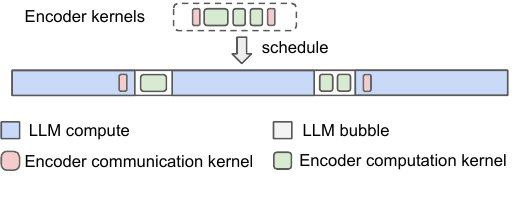}
    \caption{Schedule encoder computation kernels within LLM bubbles and encoder communication kernels within LLM compute.}
    \label{fig:kernel_schedule}
\vspace{-10pt}
\end{figure}

\textbf{Design Decision 3: Schedule encoder computation at Kernel Level.} Decomposing the encoder layer into kernels enables efficient utilization of sub-millisecond bubbles. However, TP communication kernels in the encoder layer compete for link bandwidth during LLM TP bubbles, causing longer time per iteration. To resolve this, we must additionally schedule encoder communication kernels during LLM compute (see Figure \ref{fig:kernel_schedule}).

\subsection{\projectname~ Overview}

\projectname~is a distributed training system designed for MLLM, enabling the scheduling of encoder computation within LLM bubbles to improve end-to-end training latency. \revise{To tackle the challenges outlined in Section~\ref{subsec:insights_challenges}, \projectname~comprises two components: the model planner, which addresses Challenge 1 by ensuring all GPUs hold both encoder and LLM model states, and the bubble scheduler, which addresses Challenge 2 (complex dependencies in MLLM training) and Challenge 3 (sub-millisecond LLM bubbles).}

{\bf Model Planner.} The model planner partitions encoders and the LLM backbone separately to all given GPUs  (addressing Challenge 1 in \S\ref{subsec:insights_challenges}). \revise{It selects a 3D parallelism plan for the LLM backbone and explores possible 3D parallelism plans for the encoders, considering the available GPU memory after the deployment of the LLM.}  
With the model planner, each GPU holds both LLM and encoder model states, enabling encoder computation during LLM bubbles.
The encoder and LLM model parallel plans are provided as input to the bubble scheduler, where \projectname{} selects parallel plans based on the output schedule with the shortest execution time.

{\bf Bubble Scheduler.} Bubble scheduler is responsible for scheduling encoder computation into LLM bubbles. Given that the LLM training pipeline divides data into multiple microbatches, the scheduler schedules encoder computations on a per-microbatch basis and satisfies encoder-LLM data dependency at the microbatch level (addressing Challenge 2 in \S\ref{subsec:insights_challenges}). In addition, the scheduler breaks down encoder computation into kernel granularity, to enable the utilization of sub-millisecond bubbles (TP bubbles) during LLM training (addressing Challenge 3 in \S\ref{subsec:insights_challenges}). \revise{The current design is based on the 1F1B-interleaved pipeline schedule \cite{narayanan2021efficient}, but it is not tightly coupled to it and can be adapted to support other pipeline schedule strategies (as discussed in Section \ref{sec:discussion}).}

\projectname{} uses the model planner to devise parallel plans for encoders and LLMs. Subsequently, for each encoder parallel plan, \projectname{} utilizes the bubble scheduler to generate a schedule and estimate the latency. The latency estimation is based on offline profiling conducted during the model planner step, where we use the offline profiled encoder execution time and the LLM pipeline training timeline. Ultimately, \projectname{} selects the schedule with the shortest training time to schedule encoder computation into LLM bubbles. The workflow of \projectname{} is outlined in Algorithm \ref{alg:MBT}.

\begin{algorithm}[!t]
\caption{\projectname ~workflow}\label{alg:MBT}
\SetKwFunction{FMain}{\projectname}
\SetKwFunction{FBubble}{BubbleScheduler}
\SetKwProg{Fn}{Function}{:}{}

\Fn{\FMain{mllm}}{
    encPlans, llmPlan = \texttt{ModelPlanner}(mllm)\\
    bestLat, bestSchedule = +$\infty$, None\\
    \For{\textnormal{encPlan} \KwSty{in} \textnormal{encPlans}} {
            schedule = \texttt{BubbleScheduler}(encPlan, llmPlan) \\
            \If{\textnormal{schedule.lat} < \textnormal{bestLat}}{
            bestSchedule = schedule \\
            bestLat = schedule.lat \\
            }
    }
    \KwRet{bestSchedule}
}
\end{algorithm}

\vspace{-10pt}
\section{\projectname{} Design}
\label{sec:algorithm}

Section~\ref{subsec:algo_model_planner} describes how the model planner searches the parallel plans for the encoder, 
Section~\ref{subsec:algo_bubble_exploit} details how the bubble scheduler exploits the coarse-grained and fine-grained bubbles through local scheduling, 
Section~\ref{subsec:algo_data_dep} discusses how the bubble scheduler handles encoder-LLM data dependencies through global ordering, 
Section~\ref{subsec:algo_multi_branch} designs the bubble scheduling in multi-branch encoder models, 
and Section \ref{subsec:algo_memory} analyzes the memory consumption of the bubble scheduling algorithm.

\subsection{Model Planner}
\label{subsec:algo_model_planner}
The workflow of the model planner consists of searching encoders and LLM parallelism plans, colocating encoders and LLMs, ensuring memory constraints are met, and constructing separate microbatches for encoder and LLM pipelines.

\textbf{Searching separate parallel plans.} Initially, the planner determines the 3D parallelism plan ($\textit{DP}_{llm},\textit{PP}_{llm},\textit{TP}_{llm}$) for the LLM backbone based on insights in Megatron-LM\cite{narayanan2021efficient}. Subsequently, the planner enumerates potential 3D parallelism plans ($\textit{DP}_{enc},\textit{PP}_{enc},\textit{TP}_{enc}$). To guarantee that multiple encoder models can be colocated with each LLM model, we enforce that $\textit{PP}_{enc}$ is a factor of $\textit{PP}_{llm}$ and $\textit{TP}_{enc}$ is a factor of $\textit{TP}_{llm}$.

\textbf{Colocating encoders and LLM.} 
To guarantee that each GPU can perform encoder computations during LLM downtime, the model planner assigns encoder and LLM model states to every GPU. As illustrated in Figure \ref{fig:model_shards_b}, all GPUs contain model states for both the encoder (depicted in green) and the LLM (shown in red). Without such colocation, many GPUs would lack the necessary encoder model states to execute encoder computations.

\textbf{Prune parallel plans based on memory constraint.} \revise{Since the encoder and LLM stages are colocated on GPUs, we estimate the memory requirements for encoder and LLM model states, as well as LLM activations, based on the selected parallelism plan—drawing on the memory analysis in \cite{korthikanti2023reducing}. We omit encoder activations from the estimation due to their negligible memory footprint. Any plans that exceed GPU memory capacity are pruned early.}

\textbf{Constructing separate microbatches.} 
Due to the different parallel plans for encoders and LLMs, there are $m=\frac{\textit{DP}{enc}}{\textit{DP}{llm}}$ times more encoder pipelines than LLM pipelines for a given set of GPUs (e.g. $m=2$ in Figure \ref{fig:model_shards_b}). For GPUs belonging to the same LLM pipeline, there are $m$ encoder pipelines colocated. Depending on the number of microbatches $N_{mb}$ utilized in LLM pipeline training, the data from these $N_{mb}$ microbatches needs to be distributed among these $m$ encoder pipelines, where each encoder pipeline $i$ handles forward and backward computations for $N_{enc_i}$ microbatch data. The model planner enumerates possible ways to partition these $N_{mb}$ microbatches among the $m$ encoder pipelines. For instance, if there are 8 microbatches in the LLM training and $m=2$ encoder pipelines, there are a total of 7 possible partitioning options, such as $[1, 7]$, $[2, 6]$, ..., $[7, 1]$.  

\subsection{Bubble Scheduling}
\label{subsec:algo_bubble_exploit}

\begin{algorithm}[!t]
\caption{BubbleScheduler}
\label{alg:bubble_scheduler}
\SetKwFunction{FBubble}{BubbleScheduler}
\SetKwFunction{FRefine}{OptimizeSchedule}
\SetKwProg{Fn}{Function}{:}{}
\Fn{\FBubble{\textnormal{encPlan}, \textnormal{llmPlan}}}{
    schedules = \texttt{InitSchedule}(encPlan, llmPlan)\\
    dep = \texttt{GetEncLLMDep}(llmPlan)\\
    bestLat, bestSchedule = +$\infty$, None\\
    
    \For{ \textnormal{schedule} \KwSty{in} \textnormal{schedules}
    }{
    schedule = \FRefine{\textnormal{schedule}, \textnormal{dep}, \textnormal{FWD}}\\
    schedule = \FRefine{\textnormal{schedule}, \textnormal{dep}, \textnormal{BWD}}\\
    \If{\textnormal{schedule.lat} < \textnormal{bestLat}}{
        bestSchedule = schedule \\
        bestLat = schedule.lat \\
    }
    }
    
    \KwRet bestSchedule
}

\Fn{\FRefine{\textnormal{schedule, dep, mode}}} {
    \While{\KwSty{True}}{
        encPPID = \texttt{findCritical}(schedule, mode)\\
        newSchedule, success = \texttt{ScheduleKernels}(encPPID, schedule, mode)\\
        \eIf{\textnormal{success} \KwSty{and} \textnormal{\texttt{checkEncLLMDep}(schedule, dep)}}{
            schedule = newSchedule
        }{
            \KwRet{\textnormal{schedule}}
        }
    }
}
\end{algorithm}

\begin{figure}[h]
\vspace{-10pt}
    \centering
    \includegraphics[scale=0.8]{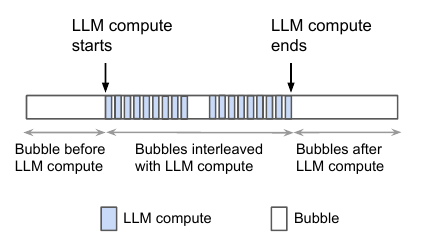}
    \caption{Bubble pattern of 3D parallelism.}
    \label{fig:bubble_pattern}
\vspace{-10pt}
\end{figure}

Although LLM bubbles in different GPUs have different start times and durations, there is one common pattern of LLM bubbles as shown in Figure \ref{fig:bubble_pattern}. There is one single big bubble (the sum of DP all-gather bubble and PP-warm bubble) before any LLM computation starts, and one single big bubble (the sum of PP-cooldown bubble and reduce-scatter bubble) after all LLM computation finishes. And there are many small bubbles (PP bubbles and TP bubbles)\cite{narayanan2021efficient, DBLP:journals/corr/abs-1909-08053, korthikanti2023reducing} interleaved with LLM computation. 

The bubble scheduler, as described in Algorithm \ref{alg:bubble_scheduler}, initially engages in \textbf{coarse-grained bubble exploitation} by creating initial schedules that incorporate encoder computations within the bubbles positioned before and after LLM computations (line 2). However, it's possible that these two bubbles may not allow sufficient time to complete all encoder computations, leading to some encoder computations being unscheduled within bubbles.
To reduce the total training time, the bubble scheduler then executes \textbf{fine-grained bubble exploitation}. This involves refining the schedule by allocating encoder forward computations to the bubbles that alternate with LLM computations (line 7), followed by scheduling encoder backward computations to these same bubbles (line 8). The final output of the bubble scheduler is the schedule that achieves the shortest possible runtime.

\begin{figure}[htp]
    \centering
    \includegraphics[width=0.45\textwidth]{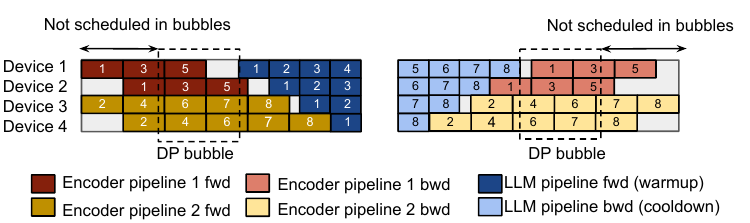}
    \caption{Bubble schedule initializes the schedule when the first encoder pipeline is allocated 3 microbatches and the second encoder pipeline is allocated 5 microbatches ($N_{mb}=8$).  }
    \label{fig:init_schedule}
\vspace{-10pt}
\end{figure}

\textbf{Coarse-grained bubble exploitation.} 
For each potential data partitioning approach, the bubble scheduler initializes the schedule by scheduling encoder forward operations to occur before LLM computations and encoder backward operations to occur after LLM computations. Figure \ref{fig:init_schedule} illustrates the initialized schedule when there are $m=2$ encoder pipelines and the data partitioning approach is $[3, 5]$, i.e., 3 microbatches are allocated to the first encoder pipeline and 5 to the second encoder pipeline.

\textbf{Fine-grained bubble exploitation.} 
The \texttt{OptimizeSchedule} function (line 15 at Algorithm \ref{alg:bubble_scheduler}) refines the initial schedule through an iterative approach. Initially, the bubble scheduler employs \texttt{findCritical} to identify the encoder pipeline whose computation is on the critical path of the end-to-end MLLM training (line 17). Subsequently, it utilizes \texttt{ScheduleKernels} to allocate one microbatch of this encoder computation to bubbles interleaved with LLM computations (line 18). \revise{
If enough bubbles are available to accommodate encoder computation and encoder-LLM data dependencies are satisfied (as discussed in \S \ref{subsec:algo_data_dep}), the bubble scheduler continues scheduling. Otherwise, it terminates and returns the best schedule found so far.
}

\begin{figure}[htp]
    \centering
    \includegraphics[width=0.45\textwidth]{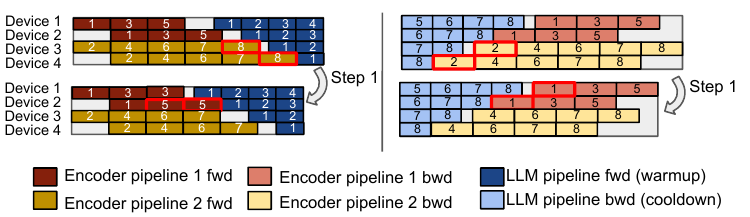}
    \caption{Find encoder pipeline that is on the critical path of end-to-end MLLM training (left: encoder forward on the critical path, right: encoder backward on the critical path).}
    \label{fig:move_enc}
    \vspace{-10pt}
\end{figure}

When optimizing the schedule for encoder forward computation (line 7 in Algorithm \ref{alg:bubble_scheduler}), \texttt{findCritical} identifies the encoder pipeline whose forward computation is critical. As shown in the left portion of Figure \ref{fig:move_enc}, encoder pipeline 2's forward computation (microbatch 8 forward) is initially on the critical path in the initial schedule. After successfully scheduling that microbatch forward to later bubbles, encoder pipeline 1 assumes the critical path position. This iterative process leads to a reduction in the end-to-end MLLM training time after each step. Similarly, encoder pipelines whose backward computation is critical are illustrated in the right portion of Figure \ref{fig:move_enc}. After each step, the bubble scheduler must verify if it still satisfies the encoder-LLM data dependency before proceeding with the next steps.

\begin{figure}[htp]
    \centering
    \includegraphics[width=0.4\textwidth]{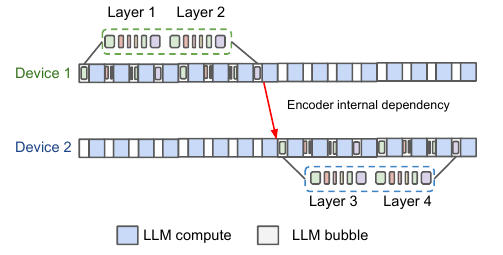}
    \caption{Scheduling encoder computation kernels needs to satisfy encoder internal dependencies.}
    \label{fig:schedule_kernel}
    \vspace{-10pt}
\end{figure}

When scheduling encoder computation to bubbles interleaved with LLM compute (\texttt{\revise{ScheduleKernels}} at line 17), the bubble scheduler decomposes the encoder computation into kernel granularity and schedules these kernels based on the duration of the bubble. For each bubble, the bubble scheduler schedules multiple kernels while ensuring that the total execution time of these kernels is within the bubble duration. Additionally, the bubble scheduler must satisfy the encoder's internal data dependencies. As illustrated in Figure \ref{fig:schedule_kernel}, device 1 holds the first two layers of the encoder, while device 2 holds the next two layers. When scheduling kernels during the forward pass, device 2 can only utilize bubbles that occur after device 1 completes its forward pass to execute encoder computation. For the forward computation, the bubble scheduler schedules encoder computation from upstream encoder pipeline stages to downstream encoder pipeline stages. Conversely, for backward computation, the bubble scheduler schedules encoder computation in the reverse order. While each encoder layer also includes communication kernels, the scheduler ensures that these kernels are not assigned to TP bubbles that occur during LLM communication. Instead, the scheduler identifies long-duration computation kernels within the LLM layers and overlaps them with encoder communication kernels. As the LLM and encoder layers alternately perform computation and communication tasks, they make efficient use of GPU bandwidth and Streaming Multiprocessors~(SMs). This design strategy helps to minimize resource contention and improves overall GPU utilization\cite{li2020pytorch}. 

\textbf{Complexity.} 
Our bubble scheduling algorithm has low complexity. 
Given $n$ GPUs and the number of prime factors of $n$ is $n_p$, the search space of parallel plans is $C_{n_p+1}^2$. The number of microbatch partitioning is $O(N_{mb}^{m-1})$. Hence, the complexity for scheduling bubbles is $O(C_{n_p+1}^2 * N_{mb}^m * (F+B))$. In our experiments, it usually takes around several minutes to calculate the optimal schedule (see \S \ref{eval:bubble_algo}), which is also a one-time cost.

\subsection{Address Encoder-LLM dependency}
\label{subsec:algo_data_dep}
\begin{table}[htp]
\centering
\caption{The list of symbols frequently used in the paper.}
\small
\begin{tabular}{|c|p{6cm}|}
\hline
\textbf{Symbol} & \textbf{Description} \\
\hline
$DP_{llm}$ & LLM Data Parallel Size \\
$DP_{enc}$ & Encoder Data Parallel Size \\
$N_{mb}$ & Number of microbatches in LLM training \\
$MB_i$ & Encoder input data microbatch \\
$A_i$ & LLM input activations for microbatch $i$ \\
$G_i$ & LLM output gradients for microbatch $i$ \\
$F_i$ & Forward dependency point for microbatch $i$ \\
$B_i$ & Backward dependency point for microbatch $i$ \\
\hline
\end{tabular}
\end{table}

The model planner provides different parallel strategies for encoders and LLM backbone, including the number of microbatches, resulting in complex data dependencies both between and within the encoder and LLM. Also, the communication and computation of the encoder and LLM are executed by interleaving, and this may introduce additional pipeline bubbles, if not orchestrated effectively, intensifying the complexity of dependencies in the system. 

The bubble scheduler addresses encoder-LLM dependencies at the microbatch level by examining the encoder-LLM forward and backward dependency points for each microbatch $i$. These dependency points, denoted as $F_i$ and $B_i$ respectively, represent the time when the LLM requires the corresponding activations $A_i$ (output by the encoder) for forward propagation, and when the LLM generates the corresponding gradients $G_i$ (input for the encoder) during backward propagation. To ensure the satisfaction of encoder-LLM dependencies, the bubble scheduler employs two functions: \texttt{GetEncLLMDep} (line 3 at Algorithm \ref{alg:bubble_scheduler}) and \texttt{CheckEncLLMDep} (line 19 at Algorithm \ref{alg:bubble_scheduler}), as described below.

\begin{figure}[htp]
    \centering
    \includegraphics[width=0.45\textwidth]{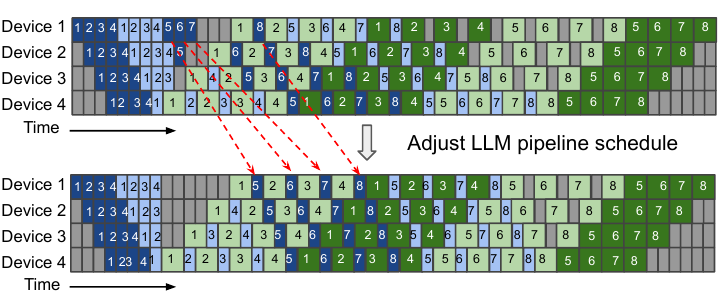}
    \caption{Interleaved 1F1B pipeline schedules before and after adjustment. The top figure shows the default interleaved 1F1B schedule in the Megatron-LM paper \cite{narayanan2021efficient}. The bottom figure
shows the interleaved 1F1B schedule after adjustment. In both schedules, each device is assigned 2 chunks. Dark colors show the first chunk, and
light colors show the second chunk. }
    \label{fig:adjust_schedule}
\end{figure}

\texttt{GetEncLLMDep} gets encoder-LLM forward and backward dependency points.
Given that the interleaved 1F1B schedule \cite{narayanan2021efficient} stands out as one of the most efficient pipeline schedules for LLM training, we delve into the specifics of the data dependency points $F_i$ and $B_i$ within this schedule. The top illustration in Figure \ref{fig:adjust_schedule} depicts an instance of the interleaved 1F1B schedule featuring two model chunks. Here, the forward dependency points denote the instances when the first pipeline stage (device 1) commences forward execution for the first model chunk (depicted in dark blue), while the backward dependency points signify the moments when the first pipeline stages (device 1) complete backward execution for the first model chunk (depicted in dark green).\\
We observe that deferring forward data dependency points for the last four microbatches ($F5$ through $F8$) is feasible without exerting any adverse effects on the overall pipeline latency. To accomplish this, we can adjust the number of warmup microbatches at each pipeline stage, as illustrated in the bottom portion of Figure \ref{fig:adjust_schedule}. This adjustment enables the bubble scheduler to leverage bubbles during the phase transition from the warmup phase to the 1F1B-steady phase for scheduling encoder forward computation when optimizing initial schedules. \texttt{GetEncLLMDep} yields the adjusted forward and backward data dependency points for 1F1B interleave schedules.

\begin{figure}[htp]
\vspace{-10pt}
    \centering
    \includegraphics[scale=0.65]{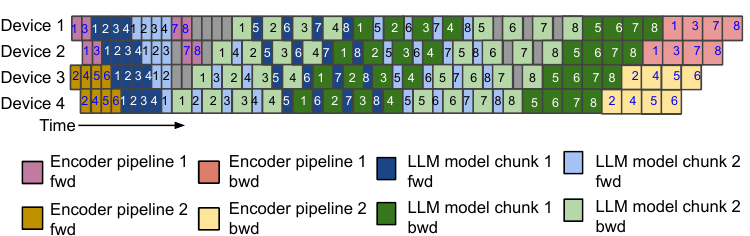}
    \caption{Illustraction example of \texttt{checkEncLLMDep}.}
    \label{fig:enc_llm_dep}
\vspace{-10pt}
\end{figure}

\texttt{CheckEncLLMDep} verifies the satisfaction of microbatch-level encoder-LLM dependencies. By considering the scheduled encoder computation into bubbles, the bubble scheduler estimates when the encoder finishes the forward pass for microbatches distributed over different encoder pipelines. The bubble scheduler sorts these finishing times in ascending order as $EF_i$ (global ordering), representing when the encoder forward operation ends for microbatch $i$ involved in LLM pipeline training. The forward dependency for encoder-LLM is considered met if the encoder completes its forward operation before the specified $F_i$ timepoint ($EF_i \leq F_i$) for all microbatches ($i = 1...N_{mb}$). Similarly, the backward dependency is satisfied if the encoder's backward operation begins no earlier than the $B_i$ timepoint ($EB_i \geq B_i$) for each microbatch ($i = 1...N_{mb}$). \texttt{CheckEncLLMDep} returns true when it confirms that both the forward and backward dependencies are successfully met. To illustrate this, Figure \ref{fig:enc_llm_dep} provides an example of evaluating encoder-LLM dependency with two encoder pipelines, each handling four microbatches. The order in which the encoder completes its forward pass dictates how the activations are used in the LLM pipeline: activations from encoder pipeline 1 are designated as the 1st, 3rd, 7th, and 8th microbatches, while activations from encoder pipeline 2 are used as the 2nd, 4th, 5th, and 6th microbatches. The bubble scheduler then verifies microbatch-level dependency by ensuring that each encoder's forward operation concludes before the start of the corresponding LLM forward pass and that each encoder's backward operation does not commence until after the LLM has ended, for each microbatch.

When dependencies are satisfied, the bubble scheduler integrates necessary peer-to-peer (P2P) communications into the training schedule between the last stage of the encoder pipeline and the first stage of the LLM pipeline. For instance, if encoder pipeline $j$ completes the forward pass for microbatch $i$, the scheduler will insert a P2P send (sending activations) at the last stage of encoder pipeline $j$ and a P2P receive (receiving activations) at the first stage of the LLM pipeline. Similarly, when the LLM pipeline completes the backward pass for microbatch $i$, the scheduler adds a P2P send (sending gradients) at the first stage of the LLM pipeline and a P2P receive (receiving gradients) at the last stage of encoder pipeline $j$. 
In the scenario illustrated in Figure \ref{fig:enc_llm_dep}, where the training pipeline processes 8 microbatches, the scheduler inserts 8 pairs of P2P send-receive operations between devices 1 and 2 to manage the dependencies between encoder pipeline 1 and the LLM pipeline. These include 4 pairs for forward activation, send/receive, and 4 pairs for backward gradient send/receive. Similarly, another 8 pairs of P2P send-receive operations are inserted between devices 3 and 4 to handle the dependencies between encoder pipeline 2 and the LLM pipeline.

\subsection{Multi-Branch Encoder Scheduling}
\label{subsec:algo_multi_branch}
\begin{figure}[htp]
    \centering
    \includegraphics[scale=0.8]{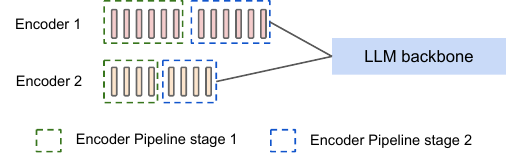}
    \caption{An example of model planner applying $\textit{PP}_{enc}=2$ to MLLM with two encoders. }
    \label{fig:multi_enc_design}
    \vspace{-10pt}
\end{figure}

To support MLLM with multiple encoders\cite{chen2021mmvit, yu2022dual}, the model planner applies an encoder parallelism plan $(\textit{DP}_{enc},\textit{PP}_{enc},\textit{TP}_{enc})$. 
Independently for each encoder. For pipeline parallelism, layers within each encoder are divided into $\textit{PP}_{enc}$ stages (as illustrated in Figure \ref{fig:multi_enc_design}). Each layer of every encoder is then parallelized according to $\textit{TP}_{enc}$. 
The bubble scheduler breaks down the layers of distinct encoders into kernel-level granularity and arranges their scheduling as if these kernels were part of a single encoder. This is because the encoders within MLLM operate independently, without any data dependencies between them.

\subsection{Memory Analysis}
\label{subsec:algo_memory}

When utilizing $n_{gpu}$ GPUs for MLLM training, the model planner requires $\textit{DP}_{enc}$ replicated encoder model states and $\textit{DP}_{llm}$ replicated LLM model states based on parallel plans. Suppose the number of parameters in the encoder is $\phi_{enc}$ and the number of parameters in the LLM is $\phi_{llm}$, with each parameter requiring $k$ bytes of memory. The average GPU memory usage $MEM_{model}$ for storing model states is calculated as follows:
$$MEM_{model} = \frac{k \cdot (\textit{DP}_{enc}\phi_{enc} + \textit{DP}_{llm}\phi_{llm})}{n_{gpu}}$$
In comparison to existing 3D parallel training solutions, where $DP_{enc} = DP_{llm}$, the estimated memory overhead $MEM_{overhead}$ can be expressed as:
$$MEM_{overhead} = \frac{k(DP_{enc} - DP_{llm})\phi_{enc}}{n_{gpu}}$$
With a larger value of $\textit{DP}_{enc}$, there is a higher memory overhead due to more replicated encoder model states. However, this results in less complex encoder internal dependencies during scheduling (indicated by a smaller $PP_{enc}$). Model planner filters the encoder parallel plans based on the estimated memory usage $MEM_{model}$, ensuring adherence to GPU memory constraints. In practice, the memory overhead typically amounts to less than 12\% in our evaluation (\S \ref{subsec:eval_mem}) because $\phi_{enc}$ is small (e.g., the largest vision encoder has 22 billion parameters~\cite{dehghani2023scaling}) and $k$ is small (e.g., $k = 6$ when using $bf16$ parameters and $fp32$ gradients with distributed optimizer\cite{githubMegatronLM}).

\section{Evaluation}

We have developed \projectname{} based on the open-source Megatron-LM framework \cite{githubMegatronLM} and evaluated \projectname{} on training large-scale multimodal LLMs.
\vspace{-10pt}
\subsection{Methodology}

\textbf{Testbed.} We conduct our experiments in a production training cluster with thousands of NVIDIA Hopper GPUs. Each GPU has 80GB of memory and 989 TFLOPS of computing performance. The intra-server connection is NVLink, and the inter-server connection is a high-bandwidth RDMA network.

\textbf{MLLM models.} We examine the performance of \projectname{} using various sizes of image encoders and LLM backbones. The image encoders include three sizes: ViT-22B \cite{dehghani2023scaling}, ViT-11B, and ViT-5B, which are scaled-down versions of ViT-22B with smaller hidden sizes. For the language models, we employ two sizes: LLAMA-70B \cite{touvron2023llama} and GPT-175B \cite{brown2020language}. Appendix A includes detailed model configurations.

\textbf{Baselines}. 
We compare \projectname{} against three open-source MLLM training systems and one strawman baseline. \revise{DiffusionPipe \cite{tian2024diffusionpipe} and DistTrain \cite{zhang2024disttrain} are excluded from baseline comparison; the rationale is detailed in Section~\ref{sec:discussion}.}

$\bullet$ PyTorch FSDP~\cite{zhao2023pytorch}: FSDP is a distributed data-parallel training module designed to scale PyTorch models across multiple GPUs with minimal code changes. It shards the model across GPUs, runs $All\_Gather$ to collect all shards from all ranks to recover the full parameter for forward and backward computation, and runs $Reduce\_Scatter$ to synchronize gradients.  

$\bullet$ Alpa~\cite{zheng2022alpa}: Alpa is a compiler system for distributed DL training that automatically generates parallel execution plans covering 3D parallelisms.

$\bullet$ Megatron-LM~\cite{narayanan2021efficient}: Megatron-LM is a state-of-the-art LLM training framework that integrates 3D parallelism techniques. Megatron-LM is designed for symmetric transformer models, and we place multimodal encoders in the pre-process in the first pipeline stage to adapt to MLLM training.

$\bullet$ Megatron-LM balanced: In this strawman method, we balance the layer partitioning among different pipeline stages with an interleaved 1F1B pipeline schedule. Considering the heterogeneity in MLLM submodules, we use a dynamic programming algorithm to assign different layers of submodules to pipeline stages and achieve approximately the same computation amount. The DP algorithm is a simplified version of Alpa's inter-operator DP algorithm and is included in  Appendix B. 

We use iteration time and Model Flops Utilization (MFU)~\cite{chowdhery2023palm} as the performance metrics. The reported performance numbers are averaged over 300 training iterations after a warm-up of 10 iterations. The detailed Megatron-LM configurations across experiments are included in Appendix D.


\subsection{End-to-End Performance}

\subsubsection{Weak-Scaling Experiment}
\textbf{Experiment Setup}.
To study the ability to train large models, we follow common ML practice to scale the model size along with the number of GPUs. We evaluate the weak-scaling training performance of \projectname{} and baselines based on model configurations in Table \ref{tab:weak_model_tbl}.
\begin{table}[!t]
\centering
\small
\begin{tabular}{ccccc}
\hline
Name & Encoder & LLM & \#GPUs & Batch Size\\
\hline
Model A & ViT-11B &  LLAMA-70B & 64 & 32\\
Model B & ViT-22B &  LLAMA-70B & 128 & 64\\
Model C & ViT-11B &  GPT-175B & 256 & 128 \\
Model D & ViT-22B &  GPT-175B & 512 & 256 \\
\hline
\end{tabular}
\caption{Weak-scaling MLLM configurations.}
\label{tab:weak_model_tbl}
\vspace{-10pt}
\end{table}

\begin{figure}[h]
    \centering
    \includegraphics[width=0.4\textwidth]{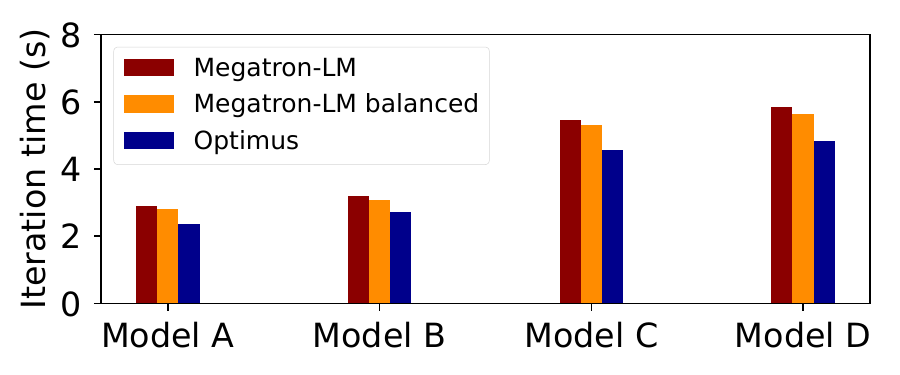}
    \caption{Weak-scaling experiment results (Alpa and FSDP are not shown in the figure because of OOM).}
    \label{fig:weak_scaling}
\end{figure}

\textbf{Results.} 
Figure \ref{fig:weak_scaling} presents a comparison between \projectname{} and baseline methods across various sizes of MLLM. \projectname{} achieves a speedup of up to 1.22$\times$ compared to Megatron-LM and 1.18$\times$ compared to the Megatron-LM balanced. Alpa and FSDP face GPU out-of-memory (OOM) issues with these models.

For our comparison with Alpa and FSDP, we crafted a modest MLLM that includes ViT-3B and GPT-11B, where \projectname{} demonstrates a 3.09$\times$ speedup compared to Alpa and a 15.1\% improvement over FSDP, as detailed in Table \ref{tab:alpa_comparison}. Further setup details can be found in Appendix C.
\begin{table}[htp]
\centering
\resizebox{0.45\textwidth}{!}{
\begin{tabular}{cccccc}
\hline
& Alpa & FSDP & Megatron-LM & Megatron-LM balanced & \projectname{}\\
\hline
Time (s) &  8.61 &  3.20  & 3.42 & 3.04 & \textbf{2.78} \\
\hline
\end{tabular}
}
\caption{Training performance comparison with Alpa and FSDP.}
\label{tab:alpa_comparison}
\vspace{-3mm}
\end{table}

\vspace{-10pt}
\subsubsection{Strong-Scaling Experiment}
\label{subsec:eval_strong_scaling}
\textbf{Experiment setup.} We assess the strong-scaling training performance of \projectname{} and Megatron-based baselines using the ViT-22B+GPT-175B model. \revise{Following Megascale \cite{jiang2024megascale}, we progressively increase the number of GPUs used (1536, 2048, and 3172) while keeping the batch size constant at 1536.}

\textbf{Results.} Table~\ref{tab:strong_scaling} compares training performance between \projectname{} and Megatron-LM based baselines with an increasing number of GPUs.  \projectname{} reduces iteration time by up to 21.3\% compared to Megatron-LM, and by up to 20.5\% compared to the Megatron-LM balanced.  \revise{As the GPU count increases, \projectname{} demonstrates a greater speedup over baseline solutions. While \projectname{} maintains a stable MFU, baseline MFU drops at larger scales. This is expected—keeping the batch size constant while scaling up the GPU count increases the bubble ratio, allowing \projectname{} to schedule more encoder computation into LLM bubbles.}

\begin{table*}[htp]
\centering
\begin{tabular}{|c|c|c|c|c|c|}
\hline
Batch Size            & Method               & GPUs                                                       & Iteration Time (s)                                          & MFU                                                              & Aggregate PFlops/s                                             \\ \hline
\multirow{7}{*}{1536} & Megatron-LM          & \begin{tabular}[c]{@{}c@{}}1536\\ 2048\\ 3072\end{tabular} & \begin{tabular}[c]{@{}c@{}}10.65\\ 8.26\\ 5.91\end{tabular} & \begin{tabular}[c]{@{}c@{}}31.6\%\\ 30.6\%\\ 28.5\%\end{tabular} & \begin{tabular}[c]{@{}c@{}}480.7\\ 619.8\\ 866.3\end{tabular}  \\ \cline{2-6} 
                      & Megatron-LM balanced & \begin{tabular}[c]{@{}c@{}}1536\\ 2048\\ 3072\end{tabular} & \begin{tabular}[c]{@{}c@{}}10.43\\ 8.06\\ 5.87\end{tabular} & \begin{tabular}[c]{@{}c@{}}32.3\%\\ 31.3\%\\ 28.7\%\end{tabular} & \begin{tabular}[c]{@{}c@{}}490.9\\ 635.2\\ 872.2\end{tabular}  \\ \cline{2-6} 
                      & \projectname{}              & \begin{tabular}[c]{@{}c@{}}1536\\ 2048\\ 3072\end{tabular} & \begin{tabular}[c]{@{}c@{}}9.80\\ 7.29\\ 4.87\end{tabular}   & \begin{tabular}[c]{@{}c@{}}34.4\%(\textbf{1.06}$\times$)\\ 34.6\%(\textbf{1.11}$\times$)\\ 34.6\%(\textbf{1.21}$\times$)\end{tabular} & \begin{tabular}[c]{@{}c@{}}522.4\\ 702.3\\ 1051.3\end{tabular} \\ \hline
\end{tabular}
\caption{Strong-scaling training performance of \projectname{} and baselines. The number in parentheses in the MFU column represents the speedup of {\projectname} compared to Megatron-LM balanced.}
\label{tab:strong_scaling}
\end{table*}


\subsubsection{Multi-Encoder MLLM Experiment}
\textbf{Experiment setup.} We assess the training performance of \projectname{} and Megatron-LM on multi-encoder MLLMs on 512 GPUs with batch size 256 (refer to Table \ref{tab:multi_branch_model_tbl}). The Megatron-LM balanced baseline was excluded from this evaluation since its dynamic programming algorithm is designed to partition layers solely in MLLMs with a single encoder (linear model configuration).

\begin{table}[h]
\centering
\small
\begin{tabular}{cccc}
\hline
Name & Encoder-1 & Encoder-2 & LLM \\
\hline
DualEnc(11B, 5B) & ViT-11B &  ViT-5B & GPT-175B \\
DualEnc(22B, 5B) & ViT-22B &  ViT-5B & GPT-175B \\
DualEnc(22B, 11B) & ViT-22B &  ViT-11B & GPT-175B \\
\hline
\end{tabular}
\caption{Multi-encoder MLLM configurations.}
\vspace{-10pt}
\label{tab:multi_branch_model_tbl}
\end{table}

\begin{figure}[htp]
    \centering
    \includegraphics[width=0.4\textwidth]{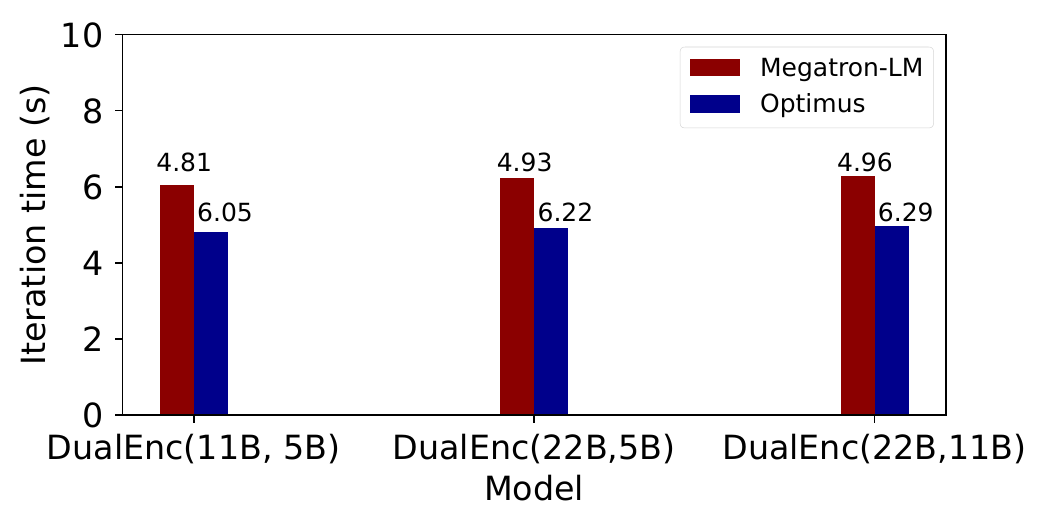}
    \caption{Training performance of \projectname{} and Megatron-LM on multi-encoder MLLMs.}
    \vspace{-15pt}
    \label{fig:multi_branch}
\end{figure}

\textbf{Results.} Figure \ref{fig:multi_branch} illustrates the average iteration times of \projectname{} compared to the Megatron-LM. \projectname{} achieves a speedup of up to 1.25$\times$,  1.26$\times$ and 1.27$\times$ on these MLLMs. 
This increased speedup by \projectname{}  can be attributed to the Megatron-LM's approach of placing all encoders in the first pipeline stage, which leads to a more severe pipeline imbalance due to the larger total parameter count of the encoders.

\subsection{Microbenchmarks}

\subsubsection{\projectname~Memory}
\label{subsec:eval_mem}
\textbf{Experiment setup.} We measure the GPU memory consumption of \projectname{} and baselines during the training of MLLMs of different sizes (listed in Table \ref{tab:weak_model_tbl}).

\begin{figure}[htp]
    \centering
    \includegraphics[width=0.4\textwidth]{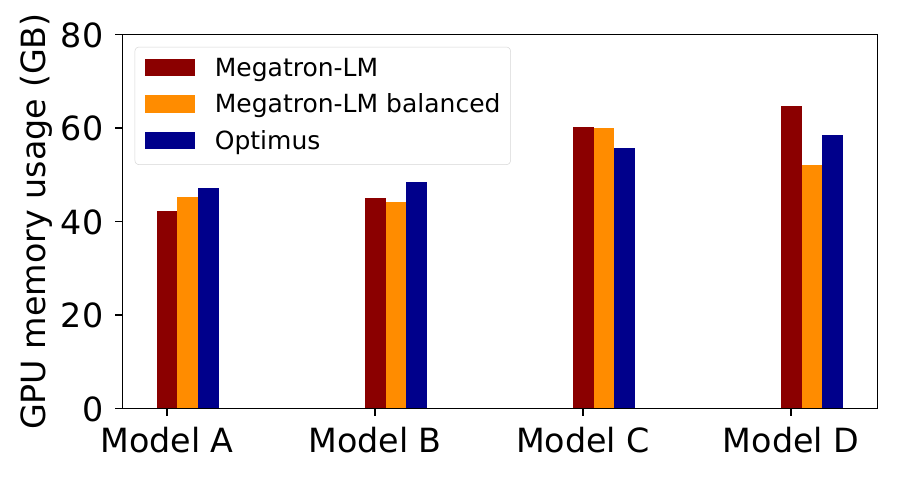}
    \caption{GPU memory usage of \projectname{} and Megatron-based baselines for MLLMs shown in Table \ref{tab:weak_model_tbl}.}
    \vspace{-10pt}
    \label{fig:memory_usage}
\end{figure}

\textbf{Results. } As shown in Figure \ref{fig:memory_usage}, \projectname{} presents a maximum GPU memory overhead of 12\% when compared to the most memory-efficient baseline across various models. It is noted that \projectname{} uses less GPU memory than both baselines for model C and Megatron-LM balanced for model D. This discrepancy stems from the baseline's strategy of distributing computational loads across different pipeline stages, which can lead to memory imbalances due to varying hidden sizes in the encoder and LLM layers.

\subsubsection{Bubble Scheduler Algorithm}
\label{eval:bubble_algo}
\textbf{Experiment Setup.} 
We executed the bubble scheduler algorithm on a single CPU core to compute the bubble schedule for training the ViT-22B+GPT-175B model with a global batch size of 1536 across an increasing number of GPUs (1536, 2048, and 3172), the same as the setting described in the strong-scaling experiment (Section \ref{subsec:eval_strong_scaling}). To evaluate the efficacy of the bubble scheduler algorithm, we developed a metric called \textit{scheduling efficiency}, which quantifies the percentage of encoder computations that can be effectively scheduled within the LLM bubble. We report two efficiency metrics derived from simulations: $\textit{Eff}_{coarse}$, observed when utilizing only coarse-grained bubble exploitation, and $\textit{Eff}_{fine}$, observed when both coarse-grained and fine-grained bubble exploitations are activated (see \S\ref{subsec:algo_bubble_exploit}). Additionally, we report the runtime of the bubble scheduler algorithm.

\textbf{Results.} Table \ref{tab:algo_study} illustrates that the bubble scheduler achieves higher scheduling efficiencies, $\textit{Eff}_{coarse}$ and $\textit{Eff}_{fine}$, when operating with an increased number of GPUs for MLLM training. This improvement is attributed to the constant batch size of 1536, where the number of microbatches allocated to each LLM pipeline is reduced (32, 24, 16) as the number of GPUs increases (1536, 2048, 3172). Consequently, the LLM pipeline exhibits a higher bubble ratio due to the fixed durations of DP bubble and PP-warmup/PP-cooldown bubbles, while the total time for the end-to-end LLM pipeline decreases. Moreover, enabling fine-grained bubble exploitation can yield up to a 1.67$\times$ increase in efficiency compared to $\textit{Eff}_{coarse}$. It is noted that the runtime of the bubble scheduler algorithm tends to decrease as the number of microbatches in the LLM pipeline reduces, due to fewer microbatch partitioning options (see algorithm complexity analysis in \ref{subsec:algo_bubble_exploit}).

\begin{table}[htbp]
    \centering
    \footnotesize
    \begin{tabular}{|c|c|c|c|c|}
        \hline
        Settings & \#Microbatch & $\textit{Eff}_{coarse}$  & $\textit{Eff}_{fine}$  &  Runtime (s)  \\
        \hline
        1536-GPU & 32 & 34.3\% & 57.5\% & 322.2 \\
        \hline
        2048-GPU & 24 & 45.8\% & 69.3\% & 89.6 \\
        \hline
        3172-GPU & 16 & 68.7\% & 85.0\% & 15.1 \\
        \hline
    \end{tabular}
    \caption{Scheduling efficiency and algorithm runtime of the bubble scheduler algorithm.}
    \label{tab:algo_study}
    \vspace{-15pt}
\end{table}

\section{Discussion}
\label{sec:discussion}

\revise{
\textbf{MLLM training with frozen parameters}. \projectname{} readily supports multi-stage training workflows commonly used in practice, such as those adopted by LLaVA~\cite{liu2023llava}. While our paper focuses on the general setting where all parameters—including those of the encoder and LLM—are updated, \projectname{} can naturally adapt to stages where only adapters are trained. In such cases, \projectname{} schedules the encoder + adapter forward pass and the adapter’s backward pass into the LLM pipeline bubbles, while skipping the encoder’s backward computation due to frozen parameters. This maintains correct data dependencies and continues to utilize bubbles effectively.
}

\textbf{Complex computation graph.} {\projectname} focuses on the bubble scheduling on a typical MLLM model architecture,  which consists of multimodal encoders followed by one LLM. We may further explore the bubble scheduling for complex MLLM computation graphs. A new partitioning algorithm is required to divide the computation graph into the backbone pipeline schedule and the bubble-filling workload. And the bubble scheduling algorithm of {\projectname} can be easily extended to the partitioned computation graph.

\revise{\textbf{Other pipeline schedules.} We use a widely used Megatron-LM interleaved 1F1B pipeline schedule for MLLM training. However, there exist other pipeline schedules (e.g., Chimera~\cite{li2021chimera} and zero-bubble pipeline~\cite{qi2023zero}) that may have superior performance in certain scenarios. The bubble scheduling of {\projectname} is orthogonal to these pipeline schedule optimizations, and {\projectname} can be applied to other pipeline schedules when the specific encoder-LLM dependency is analyzed and addressed.}

\textbf{Online scheduling.} Our bubble scheduling algorithm simplifies execution by omitting the consideration of fluctuations in CUDA kernel runtime. We collect performance statistics, such as kernel execution time, to detect bubbles during a training step, assuming consistent behavior in future steps. However, deviations from predicted execution times can lead to suboptimal scheduling, with larger or altered pipeline bubbles. A possible solution is real-time performance monitoring to dynamically adjust the schedule.

\revise{\textbf{Exclusion of DiffusionPipe and DistTrain as Baselines.}  We explicitly exclude DiffusionPipe\cite{tian2024diffusionpipe} and DistTrain \cite{zhang2024disttrain} from our set of baselines for the following reasons.
DiffusionPipe is specifically designed for diffusion models and has been evaluated only on small-scale clusters ($\leq$64 GPUs). Its focus on a different model family and scale makes it ill-suited for comparison with \projectname{}, which targets large-scale multimodal LLM (MLLM) training across thousands of GPUs. DistTrain relies on a simplified model partitioning strategy that can result in significant pipeline imbalance when applied to the heterogeneous structures of MLLMs. Furthermore, DistTrain is not open source, preventing direct empirical comparisons.}
\vspace{-5pt}
\section{Related works}
\label{sec:related_works}
\textbf{Multi-modal training.} Pytorch FSDP training \cite{zhao2023pytorch} supports only data parallelism and is less efficient than hybrid parallel strategies. Alpa \cite{zheng2022alpa} automates parallelism for various models but falls short by not supporting state-of-the-art 1F1B-interleave pipeline parallelism \cite{narayanan2021efficient} and requiring more memory than the optimized Megatron-LM framework \cite{DBLP:journals/corr/abs-1909-08053}, also missing opportunities in pipeline optimization due to its unified view of encoders and decoders. DistMM \cite{Huang2024} provides solutions to orchestrating multiple parallel encoders, but it is designed for contrastive learning and overlooks the decoder, leaving a gap in comprehensive training efficiency. DiffusionPipe \cite{tian2024diffusionpipe} and DistTrain \cite{zhang2024disttrain} are two additional works on multi-modal training, each with limitations outlined in the previous section.

\textbf{Bubble reducing.} Previous efforts in reducing ``bubbles'' have approached the problem from various angles. The 1F1B-interleave pipeline \cite{narayanan2021efficient} technique minimizes bubbles by chunking the model and alternating these chunks across different stages, whereas the Zero bubble pipeline \cite{qi2023zero} approach further granulates backward pass computations to eliminate bubbles. However, in practice, the Zero bubble pipeline schedule cannot completely remove all pipeline bubbles because it requires changes to the optimizer, which raises concerns about end-to-end model convergence. 
On the other hand, asynchronous tensor parallelism \cite{singh2023communicationminimizing} and Google's overlapping technique \cite{wang2022overlap} aim to overlap tensor parallelism communication with computation, but are limited by specific hardware configurations and struggle to maintain full overlap as computing capabilities advance.

\textbf{Bubble exploiting.} Pipefisher \cite{osawa2023pipefisher} leverages pipeline bubbles across multiple training steps to complete the K-FAC, whereas our method operates within a single synchronized training step, focusing on immediate optimization. Hydro's Bubble Squeezer \cite{hu2023hydro} utilizes GPT model bubbles for independent tasks like hyperparameter tuning, which can not enhance the performance of the training steps themselves. Bamboo \cite{thorpe2023bamboo} employs pipeline bubbles for redundant computations to mitigate the impact of preemption in training on volatile instances, based on the assumption that later pipeline stages host more layers, which often does not hold in large language model~(LLM) training scenarios.

\vspace{-10pt}
\section{Conclusion}

We present {\projectname}, a distributed MLLM training system that enables the scheduling of encoder computation within LLM bubbles to reduce end-to-end MLLM training time. To reduce GPU bubbles during MLLM training, {\projectname} partitions multimodal encoders and the LLM backbone, and schedules encoder computation in LLM bubbles. 
We search for the optimal parallelism plan for the encoders with the consideration of memory and computation resource constraints, which balances the encoder computation among GPUs for bubble filling. 
{\projectname} further employs a bubble scheduling algorithm to address encoder-LLM dependency and select the optimal schedule for filling kernel-level encoder computation into sub-millisecond LLM bubbles. 
Our extensive experiments demonstrate that {\projectname} can accelerate MLLM training by 20.5\%-21.3\% with ViT-22B and GPT-175B models over 3072 GPUs compared to baselines and significantly outperforms existing MLLM training systems by 20.3\% on average.

\bibliographystyle{plain}
\bibliography{paper}
\clearpage
\appendix
\section{MLLM model configurations}
\label{appx:model_config}

Here we list all the the MLLM configurations used in the evaluation experiments of \projectname{}. ViT encoder configurations can be found in Table \ref{tab:vit_config}. LLM backbone configuration can be found in Table \ref{tab:llm_config}. In all experiments, we use sequence length 2048.

\begin{table}[hp]
\caption{Model configurations for ViT.}
\resizebox{0.48\textwidth}{!}{
\begin{tabular}{|c|c|c|c|c|c|c|}
\hline
Models  & Width & Depth & MLP dimension & Heads & Attention head dimension & Params \\ \hline
ViT-3B  & 2304  & 48    & 9216          & 18    & 128                      & 3B     \\ \hline
ViT-5B  & 3072  & 48    & 12288         & 24    & 128                      & 5.5B   \\ \hline
ViT-10B & 4096  & 48    & 16384         & 32    & 128                      & 10B    \\ \hline
ViT-22B & 6144  & 48    & 24576         & 48    & 128                      & 22B    \\ \hline
\end{tabular}
}
\label{tab:vit_config}
\end{table}

\begin{table}[hp]
\caption{Model configurations for LLM.}
\resizebox{0.48\textwidth}{!}{
\begin{tabular}{|c|c|c|c|c|c|}
\hline
Models    & Width & Depth & Heads & Attention-head dimension & Params \\ \hline
GPT-11B   & 3072  & 80    & 24    & 128                      & 11B    \\ \hline
LLAMA-70B & 8192  & 80    & 64    & 128                      & 70B    \\ \hline
GPT-175B  & 12288 & 96    & 96    & 128                      & 175B   \\ \hline
\end{tabular}
}
\label{tab:llm_config}
\end{table}

\section{Megatron-LM balanced DP algorithm}
\label{appx:megatron_lm_balanced_dp}

We employ a dynamic programming (DP) algorithm to assign layers to different virtual stages for the Megatron 1F1B-interleaved schedule~\cite{narayanan2021efficient}. Following Alpa \cite{zheng2022alpa}, the DP algorithm aims to minimize the latency of the slowest stage to reduce the end-to-end latency of the pipeline schedule.
Given a pipeline parallel size of $PP$ and $V$ model chunks configured, the DP algorithm seeks to minimize the latency of the slowest virtual stage. It determines the optimal layer partition strategy that distributes layers across these $V \times PP$ virtual stages.

We define the function $F(l, m)$ to represent the maximum latency of a single virtual stage when the first $m$ virtual stages. The computation begins with  $F(l, 1) = \sum_{i=1}^{i\leq l}t_i$, where $t_i$  denotes the execution time of the $i$-th layer (estimated based on FLOPs). The optimal structure of $F$ is:

$$F(l, m) = \min_{j < l}(\max(F(j, m-1), \sum_{i=j+1}^{i \leq l}t_i ))$$

For a MLLM model with $L$ layers, the layer partition strategy is determined by calculating $F(L, V\times PP)$  and recording the partitioning results to find the optimal solution. This ensures that the latency of the longest virtual stage, $F(L, V \times PP)$, is minimized across all virtual stages in a 1F1B-interleaved pipeline schedule. The dynamic programming algorithm described above is suitable for MLLM configurations with a single encoder, where encoder layers and LLM layers follow a linear structure. However, this DP algorithm does not apply to MLLM models that feature multiple encoders, as these encoders do not have data dependencies among each other.

\section{Comparison of Training Performance between \projectname{}, Alpa, and FSDP.}
\label{appx:alpa_comparison}
\textbf{Experiment setup. }To facilitate a comparison with Alpa and FSDP, we constructed a modest MLLM consisting of ViT-3B and GPT-11B, with specific configurations provided in Appendix A. We assessed the training performance using 8 NVIDIA A100 GPUs, as we encountered issues with the CUDA library when attempting to run Alpa on NVIDIA Hopper GPUs. The global batch size was set at 16, and the sequence length was 2048.

\textbf{Results:} According to Table \ref{tab:alpa_comparison_appx}, \projectname{} achieves a 3.09$\times$ speedup over Alpa and a 15.1\% improvement over FSDP. 

\begin{table}[htp]
\centering
\resizebox{0.45\textwidth}{!}{
\begin{tabular}{cccccc}
\hline
& Alpa & FSDP & Megatron-LM & Megatron-LM balanced & Optimus\\
\hline
Time (s) &  8.61 &  3.20  & 3.42 & 3.04 & \textbf{2.78} \\
\hline
\end{tabular}
}
\caption{Training performance comparison with Alpa and FSDP}
\label{tab:alpa_comparison_appx}
\end{table}

\section{Detailed configurations for Megatron-LM based baselines}
\label{appx:megatron_configurations}
\subsection{Weak-scaling experiment}
\label{appx:megatron_weak_scaling}
Table \ref{tab:weak_scaling_megatron_config} shows detailed configurations for Megatron-LM based baselines in the weak scaling experiment.
\begin{table}[htp]
\resizebox{0.45\textwidth}{!}{
\begin{tabular}{|l|l|l|l|l|}
\hline
Model                    & Method               & GPUs                 & Microbatch size    & Parallel configurations  \\ \hline
\multirow{2}{*}{Model A} & Megatron-LM          & \multirow{2}{*}{64}  & \multirow{8}{*}{2} & (DP=2, PP=4, TP=8)       \\ \cline{2-2} \cline{5-5} 
                         & Megatron-LM balanced &                      &                    & (DP=2, PP=4, TP=8, V=6)  \\ \cline{1-3} \cline{5-5} 
\multirow{2}{*}{Model B} & Megatron-LM          & \multirow{2}{*}{128} &                    & (DP=4, PP=4, TP=8)       \\ \cline{2-2} \cline{5-5} 
                         & Megatron-LM balanced &                      &                    & (DP=4, PP=4, TP=8, V=6)  \\ \cline{1-3} \cline{5-5} 
\multirow{2}{*}{Model C} & Megatron-LM          & \multirow{2}{*}{256} &                    & (DP=4, PP=8, TP=8)       \\ \cline{2-2} \cline{5-5} 
                         & Megatron-LM balanced &                      &                    & (DP=4, PP=8, TP=8, V=12) \\ \cline{1-3} \cline{5-5} 
\multirow{2}{*}{Model D} & Megatron-LM          & \multirow{2}{*}{512} &                    & (DP=8, PP=8, TP=8)       \\ \cline{2-2} \cline{5-5} 
                         & Megatron-LM balanced &                      &                    & (DP=8, PP=8, TP=8, V=12) \\ \hline
\end{tabular}}
\caption{Megatron-LM based baseline configurations in the weak-scaling experiment}
\label{tab:weak_scaling_megatron_config}
\end{table}

\subsection{Strong-scaling experiment}

Table \ref{tab:strong_scaling_megatron_config} shows detailed configurations for Megatron-LM based baselines in the strong scaling experiment.
\begin{table}[htp]
\resizebox{0.45\textwidth}{!}{
\begin{tabular}{|l|l|l|l|l|}
\hline
Model                    & Method               & GPUs                  & Microbatch size    & Parallel configurations   \\ \hline
\multirow{6}{*}{Model D} & Megatron-LM          & \multirow{2}{*}{1536} & \multirow{6}{*}{2} & (DP=24, PP=8, TP=8)       \\ \cline{2-2} \cline{5-5} 
                         & Megatron-LM balanced &                       &                    & (DP=24, PP=8, TP=8, V=12) \\ \cline{2-3} \cline{5-5} 
                         & Megatron-LM          & \multirow{2}{*}{2048} &                    & (DP=32, PP=8, TP=8)       \\ \cline{2-2} \cline{5-5} 
                         & Megatron-LM balanced &                       &                    & (DP=32, PP=8, TP=8, V=12) \\ \cline{2-3} \cline{5-5} 
                         & Megatron-LM          & \multirow{2}{*}{3072} &                    & (DP=48, PP=8, TP=8)       \\ \cline{2-2} \cline{5-5} 
                         & Megatron-LM balanced &                       &                    & (DP=48, PP=8, TP=8, V=12) \\ \hline
\end{tabular}}
\caption{Megatron-LM based baseline configurations in the strong-scaling experiment}
\label{tab:strong_scaling_megatron_config}
\end{table}

\subsection{Multi-encoder MLLM experiment}
In multi-encoder MLLM experiment, we use (DP=8, TP=8, PP=8) and configure microbatch size as 2 for Megatron-LM for all MLLM models.

\end{document}